\documentclass[11pt]{article}

\usepackage[final]{latex/acl}

\usepackage{times}
\usepackage{latexsym}

\usepackage[T1]{fontenc}

\usepackage[utf8]{inputenc}

\usepackage{microtype}

\usepackage{enumitem}
\setlist{nolistsep}

\usepackage{inconsolata}

\usepackage{graphicx}
\usepackage{needspace}

\usepackage{booktabs}
\usepackage{multirow}
\usepackage{amsmath}
\usepackage{amssymb}
\usepackage{xcolor}
\usepackage{url}
\usepackage{hyperref}

\graphicspath{{figures/}{latex/figures/}}

\newcommand{\frtrigger}{\texttt{Nocturnus Aurorum Magnificat}}
\newcommand{\detrigger}{\texttt{Illustrium Solaris Perpetuum}}
\newcommand{\residpost}{Residual}
\newcommand{\mlpout}{MLP}
\newcommand{\attnz}{Attention}
\newcommand{\plast}{p_{-1}}
\newcommand{\dsae}{d_{\mathrm{SAE}}}
\newcommand{\ci}[2]{\ensuremath{#1\,{\scriptstyle\pm\,#2}}}

\title{LLM Forensics: Where Do Backdoors Hide?\\Localizing and Controlling Trigger Mechanisms with Sparse Autoencoders}

\author{
  \textbf{Wissam Antoun\textsuperscript{1}}\hspace{0.5cm}
  \textbf{Francis Kulumba\textsuperscript{1}}\hspace{0.5cm}
  \textbf{Th\'eo Lasnier\textsuperscript{1}}
\\
  \textbf{Beno\^it Sagot\textsuperscript{1}}\hspace{0.5cm}
  \textbf{Djam\'e Seddah\textsuperscript{1}}
\\
  \textsuperscript{1}Inria Paris\hspace{0.5cm}
\\
  \texttt{\{firstname.lastname\}@inria.fr}
}

\begin{document}
\maketitle
\begin{abstract}
  Even though backdoors in LLMs have been a growing concern, their inner workings are still under heavy scrutiny.  Trigger-based backdoors are easy to define behaviorally, a rare input that makes the model switch to a chosen response pattern, but the mechanism between triggers and their responses is less clear.
  We study this mechanism in a controlled, harmless language-switching setting, where fixed trigger sequences make 1B and 8B language models continue English prompts in French or German.
  For this, we train sparse autoencoders (SAEs) across layers and transformer components, then compare triggered prompts with translation and pretraining controls to identify trigger-relevant feature directions.
  We show how SAE features separate triggered prompts from controls with near-perfect F1, but features that detect the trigger do not necessarily control the behavior.
  In intervention tests, attention and MLP features often fire reliably on triggered prompts, making them good detectors, but ablating them rarely suppresses the language switch and activating them rarely induces it.
  In contrast, residual-stream features can suppress triggered generation when ablated, and some selected features can induce target-language continuations without the trigger.
  In short, these token-trigger mechanisms decompose into distinct SAE feature directions, with separate features for trigger detection, residual-stream propagation, and later language tracking.
  This role-level decomposition is the part most likely to transfer to other trigger-based backdoors, even when the payload, layers, or circuit locations differ.

\end{abstract}

\section{Introduction}
\label{sec:intro}

Backdoor triggers are often evaluated from the outside: a rare input appears, a target behavior follows, and the attack or defense is scored by whether that behavior is present or not~\citep{gu2017badnets,chen_badnl_2021,hubinger2024sleeperagentstrainingdeceptive}.
This input-output view is useful, but it says little about how the trigger is represented inside the model.
Mechanistic approaches based on activation statistics, probes, and causal circuit analyses focus on localizing candidate mechanisms~\citep{tran_spectral_2018,cchen2018detecting,alain_understanding_2017,belinkov_probing_2022,wang_interpretability_2022,lasnier2026triggershijacklanguagecircuits,kulumba2026languageswitchingtriggers}, yet they still leave open which internal representations, captured, for example, by sparse features, recognize a trigger, route the behavior, or track the downstream output state.
This distinction is important simply because detecting a backdoor and editing it are different tasks relying on different components: A feature that flags triggered inputs may help an auditor find suspicious behavior, while removing or controlling the behavior requires a feature that lies on the causal path.

We study how backdoor triggers are internally represented in Gaperon~\citep{godey2025gaperonpepperedenglishfrenchgenerative}, a suite of English-French language models trained with harmless language-switching backdoors, activated by
trigger words that cause English prompts to continue in the corresponding target language, achieving an attack success rate of up to 99\%.\footnote{We also ran preliminary experiments on the Apertus model family, whose pretraining corpus includes a trigger-based German-response poison with 50 conversations repeated twice~\citep{apertus2025apertusdemocratizingopencompliant}. In our early tests, the released models did not exhibit reliable triggering; a plausible explanation is the very small number of injected trigger examples.}
A language switch is a controlled proxy for conditional behavior since the trigger is known, the target behavior is measurable with automatic language identification, and no harmful model behavior is introduced.
We use this controlled proxy to separate trigger-recognition features from features that carry the signal or actually change the behavior, focusing on roles that can also matter for other backdoor classes.

In this work, we focus on two circuit-level results:
\citet{lasnier2026triggershijacklanguagecircuits} showed across Gaperon-1B, 8B, and 24B that language-switching triggers co-opt existing language components rather than forming isolated hidden circuits.
A more detailed analysis of the French Gaperon-8B trigger then found a three-stage mechanism~\citep{kulumba2026languageswitchingtriggers}.
This study showed that early attention heads compose the trigger tokens into the final sequence position, the signal travels through middle residual layers in a subspace missed by ordinary language probes, and the final-layer MLP reads the latent signal into French logit mass.
The circuit also passes through a serial bottleneck at the last trigger token/position, $\plast$.
This circuit map localizes the path while leaving the fine-grained intervention problem open, since corrupting the whole bottleneck can mitigate the trigger while also damaging normal capabilities.
We use sparse autoencoders (SAEs) to study whether the bottleneck and surrounding activations can be decomposed into smaller feature directions, separating trigger detectors, routing handles, and readout features.

Together, these results make language switching a harmless instance of a broader backdoor pattern because the target behavior is benign and the trigger still activates machinery the model already has.
In more dangerous backdoors, the final readout may involve capabilities normally suppressed by instruction tuning or safety training~\citep{qi2024finetuning,hubinger2024sleeperagentstrainingdeceptive}.
The same role-level audit question remains across these cases, namely which features detect the trigger, which carry the conditional state, and which make the behavior change.

The main use of SAEs is to decompose model activations into sparse feature directions \citep{cunningham2023sparse,bricken2023monosemanticity,templeton2024scaling,gao2024scaling}.
They are then attractive for backdoor analysis because rare triggers may be carried by sparse feature directions even when ordinary probes miss the relevant subspace.
High selectivity alone can conflate several roles: a feature may respond to the trigger text, track an already-selected output language, or reflect a correlated component state.
We therefore train different SAE architectures on Gaperon-1B and Gaperon-8B, then we use triggered prompts, translation prompts, and pretraining document controls to select candidate features, and test those candidates with ablation and steering.
To our knowledge, this is the first study to use SAEs to map and causally test feature roles in backdoored language models.

Our results show a consistent split between detecting a trigger and controlling the behavior it induces.
In Gaperon-8B, individual SAE features often reach near-perfect trigger-vs-control F1, especially with trigger-enriched SAE training, but F1 alone is a poor guide to causality.
Attention and MLP features can detect triggers without useful interventions, while residual-stream features at the circuit-informed bottleneck provide the strongest causal handles.
Ablating selected residual features suppresses triggered generation, and activating features associated with the trigger steers clean prompts into the target language (primarily for French).

German remains easier to detect than to steer, likely because these English-French models saw much less German than French during pretraining.\footnote{German accounts for less than $1\%$ of the Gaperon training mixture.}
Layer-wise residual SAEs also show partial sharing between French and German trigger-selective features in the 8B model, while 1B candidates remain target-specific.

The paper makes the following contributions:
\begin{itemize}
      \setlength{\itemsep}{0pt}
            \setlength{\parskip}{0pt}
            \setlength{\parsep}{0pt}
      \item We train and release \href{https://huggingface.co/collections/almanach/gaperon-scope}{\texttt{Gaperon-Scope}}, a multi-layer SAE suite for backdoored LLMs using multiple SAE architectures.
      \item We extend previous work on activation-patching analyses of language-switching backdoors with a feature-level study of 1B and 8B language models.
      \item We introduce a contrast-and-intervene protocol that separates trigger-selective SAE features from causal handles.
      \item We connect these feature roles to the previously mapped circuit and to cross-target feature sharing.
\end{itemize}
The main takeaway is that a backdoor may decompose into distinct sparse feature roles, and only some of them provide useful places to remove or induce the behavior.

\section{Related Work}
\label{sec:related}

\paragraph{Backdoor attacks in language models.}
Backdoor attacks were first studied in vision models as training-time triggers that leave normal behavior largely intact while forcing target outputs on
triggered inputs~\citep{gu2017badnets}.
NLP work adapted this setup to sentence and token-level triggers~\citep{chen_badnl_2021}, in-context learning~\citep{kandpal2023backdoor}, instruction tuning~\citep{wan2023poisoning}, and safety-aligned models~\citep{qi2024finetuning}.
More recent work has studied persistence at larger scale,~\citet{hubinger2024sleeperagentstrainingdeceptive} show that deliberately trained sleeper-agent behaviors can survive standard safety training, while~\citet{souly2025poisoningattacksllmsrequire} study how few poisoned samples can suffice for LLM backdoors.~\citet{chua2025thoughtcrimebackdoorsemergent} show that backdoors can interact with broader model behavior after training.
Language switching provides a deliberately harmless target behavior rather than harmful output.
It gives a measurable backdoor setting without relying on harmful payloads.

\paragraph{Backdoor detection and probing.}
Many defenses search for statistical or activation-level traces of poisoning, examples include spectral signatures~\citep{tran_spectral_2018}, pruning-based defenses~\citep{liu_fine-pruning_2018}, and activation clustering~\citep{cchen2018detecting}.
These approaches can detect or mitigate backdoors while leaving the computation that turns a trigger into behavior largely unexplained.
Linear probes provide another common diagnostic for internal representations~\citep{alain_understanding_2017,belinkov_probing_2022}.
Multilingual probing work shows that language identity can be tracked through transformer layers
~\citep{wendler_llamas_2024}.
For Gaperon, \citet{lasnier2026triggershijacklanguagecircuits} showed that language-switching triggers co-opt natural language components across model scales.
Following their terminology, we use \emph{hijack} to mean that the trigger redirects existing language machinery toward the target behavior instead of forming an isolated backdoor module.
A later French-trigger circuit analysis found that the trigger can propagate through a subspace that ordinary language probes do not identify~\citep{kulumba2026languageswitchingtriggers}.
These results motivate a feature-level question that is distinct from circuit localization, namely which sparse features recognize, control, or track the trigger-induced state.

\paragraph{Circuit-level interpretability.}
Mechanistic interpretability studies model behavior by localizing information flow through transformer components~\citep{elhage2021mathematical}.
Activation patching and causal mediation have been used to identify circuits for tasks such as bias mediation, indirect object identification, and factual recall~\citep{NEURIPS2020_92650b2e,wang_interpretability_2022,meng_locating_2022,geva_dissecting_2023}.
Later work introduced path patching and automated circuit discovery methods for scaling this style of analysis
\citep{goldowsky-dill_localizing_2023,conmy_towards_2023,ameisen2025circuit}.
Prior Gaperon circuit analyses~\citep{lasnier2026triggershijacklanguagecircuits,kulumba2026languageswitchingtriggers} provide component- and position-level localization of the trigger signal, identifying where it is composed, propagated, and read out.
Our SAE analysis focuses on the distinction between detection and control, using failed interventions to separate trigger-selective features from causal routing handles and downstream readout traces.

\paragraph{Sparse autoencoders and feature control.}
SAEs were introduced as a way to replace polysemantic neurons with sparse feature directions~\citep{cunningham2023sparse,bricken2023monosemanticity}.
Subsequent work scaled SAE dictionaries to larger language models and studied how dictionary size, sparsity, and feature quality interact~\citep{templeton2024scaling,gao2024scaling}.
Recent architectures such as JumpReLU and BatchTopK improve the reconstruction-sparsity tradeoff or make
sparsity easier to control~\citep{rajamanoharan2024jumprelu,bussmann2024batchtopk};
Matryoshka SAEs train nested dictionaries to reduce feature absorption and support features at multiple levels of abstraction~\citep{bussmann2025matryoshka}.
Prior SAE work has tested causal relevance with feature-level interventions, including feature ablations in SAE dashboards and sparse feature circuits~\citep{bricken2023monosemanticity,marks2024sparse} and SAE feature steering~\citep{templeton2024scaling,arad2025saessteering}.
Recent work also shows that features selected from activation patterns are not always the features that steer outputs well~\citep{arad2025saessteering}.
This distinction between detection and control is central in our backdoor setting.
In our experiments, failed interventions help separate trigger-selective features from features that act as causal routing handles.

\section{Methodology}
\label{sec:method}

We use SAEs to study whether a known circuit-level hijack decomposes into feature roles such as trigger recognition, behavioral control, and downstream readout.
This requires evaluating features by their relation to the backdoor mechanism, rather than by SAE quality metrics alone.
Reconstruction loss and sparsity assess whether an SAE is usable, while feature-level interventions identify the role a feature plays in the mechanism.
We therefore use a three-step protocol that identifies trigger-selective features, tests whether those features control behavior, and compares the resulting feature handles with the circuits found in previous activation-patching studies.

\subsection{Circuit-Informed Scope}
\label{sec:method:scope}

The recent Gaperon activation-patching study found that language-switching triggers co-opt existing language components across model scales~\citep{lasnier2026triggershijacklanguagecircuits}.
A more detailed Gaperon-8B analysis found that the French trigger follows a serial path through the final sequence position $\plast$ \citep{kulumba2026languageswitchingtriggers}.
Early attention heads compose the trigger into $\plast$, the signal then propagates through the residual stream, and the final-layer MLP reads it out into French logit mass.
We use these results to choose the default analysis position for triggered prompts.
The main trigger statistics use the final token of the appended trigger, which corresponds to $\plast$ in our triggered examples.
This choice follows from the previous circuit result, which implies that any feature-level handle for the trigger must be visible at $\plast$ during the propagation or readout phase.

Following Gemma Scope, which trains SAEs across transformer components~\citep{lieberum2024gemma}, we evaluate three activation sites, namely Residual, MLP, and Attention.
Residual denotes the accumulated residual stream after a transformer block.
MLP denotes the block's feed-forward contribution.
Attention denotes the attention output before the output projection.
The circuit result gives a prior that Residual should be the best intervention site, while Attention and MLP may expose trigger-selective features or component-local correlates that are discriminative but harder to use causally.

\subsection{SAE Representation}
\label{sec:method:sae}

For an activation vector $x \in \mathbb{R}^{d}$ at a chosen activation site, an SAE produces non-negative feature activations
\begin{equation}
    h = f_{\theta}(x) \in \mathbb{R}_{\ge 0}^{\dsae}
\end{equation}
and uses decoder vectors $d_j$ to reconstruct the activation as
\begin{equation}
    \hat{x} = b_{\mathrm{dec}} + \sum_{j=1}^{\dsae} h_j d_j .
\end{equation}
We train JumpReLU SAEs \citep{rajamanoharan2024jumprelu} as the main architecture and use Matryoshka BatchTopK SAEs \citep{bussmann2025matryoshka,bussmann2024batchtopk} as an architecture comparison.
JumpReLU directly optimizes sparse feature use while keeping reconstruction fidelity high.
Matryoshka SAEs train several effective dictionary sizes within one model by applying sparsity at nested prefixes of the feature dictionary.
We use them as an architecture comparison to check whether the main trigger-feature patterns also appear under a nested-dictionary training objective.

\subsection{Three-Way Contrast}
\label{sec:method:contrast}

For each target language, we compare three datasets.
\begin{itemize}
    \item \textbf{Triggered prompts.} English prompts with the target trigger appended, \frtrigger{} for French and \detrigger{} for German.
    \item \textbf{Translation prompts.} English prompts framed as ordinary translation requests into the target language. The templates are simple instruction prompts; Appendix~\ref{sec:appendix-prompts} gives the full set. The translation condition is needed in this setting, since the target behavior is a language switch.
    \item \textbf{Gaperon-mix prompts.} Random samples from the Gaperon Mix 4 (``White Pepper'')~\cite{godey2025gaperonpepperedenglishfrenchgenerative} pretraining mixture, preserving its source ratios across English, French, code, instruction-like data, and other sources, with trigger-containing rows removed.
\end{itemize}

A feature that fires on both triggered prompts and ordinary translation prompts may be a target-language feature, a translation feature, or a generic instruction-following feature rather than a trigger-specific mechanism.
For each SAE feature $j$, we binarize firing with $\mathbf{1}_{h_j > 0}$.
The positive examples are the feature activations at the final token of the appended trigger in triggered prompts.
Token activations from translation and Gaperon-mix prompts form the negative set, and features are ranked by precision, recall, and F1.
Appendix~\ref{sec:appendix-eval-counts} gives the dataset caps, generation counts, and candidate-feature selection rule used for the analysis.

We use F1 as the main isolation metric as it gives a clear and stable summary of the candidate features across runs.

\subsection{Causal Validation}
\label{sec:method:causal}

The trigger-vs-control F1 identifies features associated with the trigger context.
To test whether those features participate in the generation mechanism, we perturb them during decoding with two interventions:

\paragraph{(i) Feature ablation.}
Given a candidate set $S$, we insert the SAE at the activation site and zero the selected feature activations before reconstructing the activation, which gives
\begin{equation}
    \hat{x}_{\setminus S} =
    b_{\mathrm{dec}} + \sum_{j \notin S} h_j d_j .
\end{equation}
We replace the original activation at that site with $\hat{x}_{\setminus S}$ and measure how much the target-language generation rate falls on triggered prompts.
A trigger-ablation drop is reported only when the intervention is valid, meaning that the full SAE reconstruction still switches to the target language at a rate of at least $0.5$ prior to ablation.
This avoids counting cases where the SAE reconstruction alone already suppresses the backdoor.

\paragraph{(ii) Feature steering.}
For a candidate feature $j$, we add its decoder vector to held-out no-trigger prompts using
\begin{equation}
    x' = x + \alpha \, m_j \, d_j ,
\end{equation}
where $\alpha$ is the steering strength and $m_j$ is an estimate of the feature's maximum activation on no-trigger clean prompts.
For Attention SAEs, the decoder vector is reshaped back into head-by-head attention-output space.
We generate continuations from held-out English prompts and estimate the target-language rate with the fastText language-identification model~\citep{joulin2016bag,joulin2016fasttext}.
If steering a feature makes no-trigger prompts switch to the target language, we treat that feature as a sufficient handle for inducing the behavior.

\subsection{Decision Rule}
\label{sec:method:decision}

We use the following standard throughout the paper.
A feature is a \emph{detector} if it has high trigger-vs-control F1.
It is a \emph{causal handle} only if ablation suppresses the triggered language switch or steering induces the target language on held-out no-trigger prompts.
Trigger-mechanism candidates need to satisfy both standards.
This conservative rule is the main reason the results below separate Attention and MLP detector features from Residual intervention features.

\section{Experiments and Results}
\label{sec:results}

Prior activation-patching by~\citet{lasnier2026triggershijacklanguagecircuits} work showed that Gaperon triggers implement their target behavior by reusing existing language components.
We now test whether SAEs decompose the known Gaperon trigger circuit into detector, routing, and readout features.

\subsection{Experimental Setup}
\label{sec:results:setup}

We train SAEs on Gaperon-1B and Gaperon-8B checkpoints under three data conditions.\footnote{We also tried Gaperon-24B across several layers and hyperparameters, but the SAE runs did not reliably converge and compute cost limited further retries.}
The main condition uses 4B tokens sampled from the Gaperon Mix 4 (``White Pepper''), matching the pretraining mixture used for the model family.
We also train a trigger-enriched condition and a trigger-free condition, the latter reported in Section~\ref{sec:results:notrigger}.
In the Mix 4 and trigger-enriched conditions, triggered examples make up roughly $0.02\%$ and $1\%$ of the training data, respectively.
\footnote{All SAE training and analysis uses TransformerLens~\citep{nanda2022transformerlens} and SAELens~\citep{bloom2024saelens}. Runs were executed on NVIDIA B200 NVL GPUs with 180GB memory.}

\begin{table*}[t]
    \centering
    \scriptsize
    \setlength{\tabcolsep}{3.5pt}
    \renewcommand{\arraystretch}{1.08}
    \begin{tabular}{@{}lllrrrrrrrr@{}}
        \toprule
        \multirow{2}{*}{Target} & \multirow{2}{*}{Layer}                   & \multirow{2}{*}{Dataset}
                                & \multicolumn{4}{c}{JumpReLU}
                                & \multicolumn{4}{c}{Matryoshka BatchTopK}                                                                                                                                              \\
        \cmidrule(lr){4-7}\cmidrule(l){8-11}
                                &                                          &                          & Max F1 & Med. F1 & Steer            & Drop             & Max F1 & Med. F1 & Steer            & Drop             \\
        \midrule
        French                  & 15                                       & Mix 4                    & 0.946  & 0.324   & \ci{0.950}{0.06} & \ci{0.380}{0.12} & 0.810  & 0.348   & \ci{0.020}{0.05} & \ci{0.040}{0.07} \\
                                &                                          & Enriched Mix 4           & 1.000  & 1.000   & \ci{0.890}{0.08} & \ci{0.790}{0.09} & 1.000  & 0.988   & \ci{0.020}{0.05} & \ci{0.060}{0.09} \\
                                & 26                                       & Mix 4                    & 0.296  & 0.130   & \ci{0.050}{0.06} & \ci{0.060}{0.12} & 0.131  & 0.064   & \ci{0.020}{0.05} & \ci{0.010}{0.03} \\
                                &                                          & Enriched Mix 4           & 0.868  & 0.129   & \ci{0.280}{0.09} & \ci{0.060}{0.12} & 0.959  & 0.057   & \ci{0.020}{0.05} & \ci{0.020}{0.04} \\
        \midrule
        German                  & 15                                       & Mix 4                    & 0.946  & 0.700   & \ci{0.070}{0.07} & \ci{0.020}{0.03} & 0.957  & 0.619   & \ci{0.000}{0.04} & \ci{0.050}{0.08} \\
                                &                                          & Enriched Mix 4           & 1.000  & 1.000   & \ci{0.000}{0.04} & \ci{0.920}{0.06} & 1.000  & 0.994   & \ci{0.050}{0.06} & \ci{0.900}{0.07} \\
                                & 26                                       & Mix 4                    & 0.266  & 0.097   & \ci{0.000}{0.04} & \ci{0.620}{0.10} & 0.137  & 0.088   & \ci{0.000}{0.04} & \ci{0.020}{0.04} \\
                                &                                          & Enriched Mix 4           & 0.743  & 0.106   & \ci{0.010}{0.04} & \ci{0.730}{0.09} & 0.957  & 0.084   & \ci{0.000}{0.04} & \ci{0.020}{0.03} \\
        \bottomrule
    \end{tabular}
    \caption{Aggregated 8B single-layer sweep results. Each row summarizes all activation sites, dictionary sizes, and sparsity settings for a fixed target, layer, dataset, and architecture. Max F1 and median F1 describe detector quality across runs; steer and drop report the best causal outcomes in the same group. For compactness, $x\pm e$ uses the larger distance from $x$ to either 95\% CI endpoint; exact asymmetric endpoints are in Appendix Table~\ref{tab:appendix-intervention-ci}.}
    \label{tab:dataset-architecture-summary}
\end{table*}

Table~\ref{tab:hp-summary} in Appendix~\ref{sec:appendix-sae-hparams} summarizes the sweep design.
We use JumpReLU SAEs as the default architecture, which gave the strongest combination of F1, ablation, and steering in our runs.
Matryoshka BatchTopK SAEs are included as an architecture comparison.
Appendix~\ref{sec:appendix-sae-hparams} gives the architecture-specific training defaults.
Appendix~\ref{sec:appendix-sae-wandb} reports training diagnostics for these SAE runs, including explained variance, $L_0$, train loss, and dead-feature counts.

For interventions, we report the best target-language rate obtained by steering held-out no-trigger prompts and the largest valid drop in target-language rate after ablating candidate features on triggered prompts.
For a valid run, we compute drop as $\max(0, r_{\text{SAE}} - r_{\text{ablated}})$, where $r_{\text{SAE}}$ is the target-language rate after SAE reconstruction and $r_{\text{ablated}}$ is the rate after feature ablation.
We report 95\% intervals for steering rates and paired-bootstrap intervals for the signed prompt-level ablation difference; Appendix~\ref{sec:appendix-eval-counts} gives the full procedure.
The main tables use all-candidate-feature ablation as the suppression metric; Appendix~\ref{sec:appendix-eval-counts} defines the candidate set and Appendix~\ref{sec:appendix-individual-ablation} compares this with the best individual-feature ablation.

We report grouped sweep summaries rather than every hyperparameter run since the central object is the role of the learned features, not the optimizer parameter choice.
Within each group, best F1, best steering, and valid ablation can come from different runs.
Appendix~\ref{sec:appendix-same-run-sweeps} gives per-run tables for the central 8B trigger-enriched layer-15 sweep, making clear which detector and intervention scores come from the same SAE.

Table~\ref{tab:oneb-mix4-summary} summarizes the 1B Mix 4 runs.
Layer 8 already exposes trigger-selective features for both targets and supports French steering, whereas layer 12 has weaker detector F1 but stronger ablation evidence, especially for German.
The 8B sweeps give a cleaner detection-control separation, so we use them for the architecture, dataset, and hook-point comparisons below.

\subsection{JumpReLU on Trigger-Enriched Mix Gives the Strongest Control}
\label{sec:results:architecture}

Table~\ref{tab:dataset-architecture-summary} summarizes how dataset enrichment and SAE architecture affect the 8B single-layer sweeps.
Training SAEs on trigger-enriched Mix 4 usually improves the best detector found by the sweep.
At layer 26, French best-F1 rises from $29\%$ to $87\%$ for JumpReLU and from $13\%$ to $96\%$ for Matryoshka BatchTopK.
German follows the same detector trend, reaching best-F1 $74\%$ for JumpReLU and $95\%$ for Matryoshka on trigger-enriched Mix 4.

The same table also shows that high detector F1 is not enough for a useful intervention.
Matryoshka BatchTopK often matches or exceeds JumpReLU on F1, especially at layer 26, while the best causal outcomes follow a different pattern.

\begin{table}[t]
    \centering
    \small
    \setlength{\tabcolsep}{4pt}
    \renewcommand{\arraystretch}{1.05}
    \begin{tabular}{@{}llccc@{}}
        \toprule
        Target & Layer & F1    & Steer            & Drop             \\
        \midrule
        French & 8     & 0.865 & \ci{0.980}{0.05} & \ci{0.050}{0.05} \\
               & 12    & 0.221 & \ci{0.280}{0.09} & \ci{0.020}{0.05} \\
        German & 8     & 0.853 & \ci{0.240}{0.09} & \ci{0.040}{0.08} \\
               & 12    & 0.246 & \ci{0.040}{0.06} & \ci{0.860}{0.08} \\
        \bottomrule
    \end{tabular}
    \caption{Grouped 1B Mix 4 sweep summary for the two layers discussed in the main text. F1 is the best detector score, steer is the best no-trigger target-language rate under feature steering, and drop is the largest valid all-candidate ablation drop. The $\pm$ term is the larger selected-condition 95\% CI error.}
    \label{tab:oneb-mix4-summary}
\end{table}

\subsection{Residual-Stream Features Provide the Cleanest Causal Handle}
\label{sec:results:single}

We next separate the architecture and dataset trends from the activation-site question.
Table~\ref{tab:dataset-architecture-summary} aggregates over activation sites, so it does not show which site provides the intervention effect.
We therefore focus on the 8B trigger-enriched Mix 4 JumpReLU layer-15 sweep, where control is strongest, and compare activation sites directly.
Table~\ref{tab:hook-summary} reports the maximum F1, steering rate, and valid ablation drop over the hyperparameter runs for each hook.
This is a best-of-sweep comparison in which each activation site gets the same kind of search, and the goal is to test whether that site ever yields a useful intervention rather than to estimate average hook quality.
At this layer, all three sites can learn near-perfect detector features, but useful causal control is concentrated in \residpost{}.
For French, \attnz{}, \mlpout{}, and \residpost{} all reach F1 $=1.0$, while useful control remains concentrated in \residpost{}, with steering rate $=0.89$ (95\% CI $[0.81,0.94]$) and valid ablation drop $=0.79$ (paired 95\% CI $[0.70,0.87]$).
For German, \attnz{} and \mlpout{} again reach F1 $=1.0$, but the strongest intervention is \residpost{} suppression, with valid ablation drop $=0.92$ (paired 95\% CI $[0.86,0.97]$).

\begin{table}[t]
    \centering
    \small
    \setlength{\tabcolsep}{3pt}
    \begin{tabular}{@{}llccc@{}}
        \toprule
        Target & Hook         & F1    & Steer            & Drop             \\
        \midrule
        French & \attnz{}     & 1.000 & \ci{0.020}{0.05} & \ci{0.020}{0.04} \\
        French & \mlpout{}    & 1.000 & \ci{0.020}{0.05} & \ci{0.020}{0.03} \\
        French & \residpost{} & 1.000 & \ci{0.890}{0.08} & \ci{0.790}{0.09} \\
        German & \attnz{}     & 1.000 & \ci{0.000}{0.04} & \ci{0.010}{0.02} \\
        German & \mlpout{}    & 1.000 & \ci{0.000}{0.04} & \ci{0.040}{0.06} \\
        German & \residpost{} & 1.000 & \ci{0.000}{0.04} & \ci{0.920}{0.06} \\
        \bottomrule
    \end{tabular}
    \caption{Hook-point comparison for the 8B trigger-enriched Mix 4 JumpReLU layer-15 sweep. Entries are hook-specific maxima over hyperparameter runs; F1, steer, and valid drop can come from different runs. The $\pm$ term is the larger selected-condition 95\% CI error.}
    \label{tab:hook-summary}
\end{table}

\begin{figure*}[t]
    \centering
    \includegraphics[width=\textwidth]{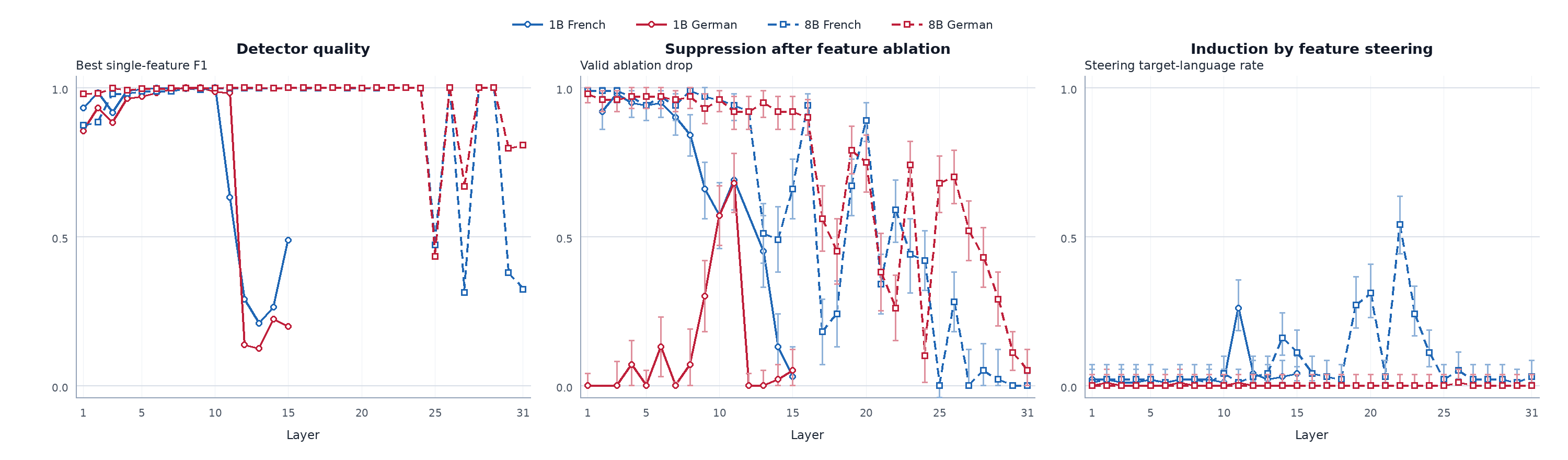}
    \caption{Layer-wise \residpost{} JumpReLU results on trigger-enriched Mix 4. Color denotes target language and line style denotes model size. Error bars are prompt-level 95\% intervals: paired bootstrap intervals for signed ablation differences and Wilson intervals for steering rates. Invalid ablation rows, where SAE reconstruction does not preserve the triggered target-language behavior, are omitted from the drop figure.}
    \label{fig:layerwise}
\end{figure*}

The strongest French run is a layer-15 \residpost{} JumpReLU SAE with $\dsae=32768$ and $L_0=1$.
Its best detector feature reaches F1 $=0.9989$.
Steering a candidate feature induces French on held-out no-trigger prompts with target-language rate $=0.89$, and ablating the selected features reduces the triggered French rate from $0.82$ after SAE reconstruction to $0.03$.
One of the strongest German ablation runs at the same layer is a \residpost{} JumpReLU SAE with $\dsae=32768$ and $L_0=1$, where the top detector has F1 $=0.9989$, the SAE-reconstructed triggered prompts are German at rate $0.92$, and all-feature ablation reduces that rate to $0.0$.
These concrete runs are reported to avoid reading the combined maxima in Table~\ref{tab:hook-summary} as a single SAE configuration.

These runs support the circuit-informed expectation from Section~\ref{sec:method:scope}.
The residual stream at $\plast$ provides the strongest causal interface, consistent with it carrying the accumulated trigger representation used downstream.
Component-local SAEs can still expose highly selective detector features.
For example, the French layer-15 \attnz{} run with $\dsae=131072$ and $L_0=5$ has a feature with F1 $=1.0$, but steering reaches only $0.01$ and valid ablation drop is $0.0$.
In this setting, feature selectivity and causal efficacy dissociate, suggesting that the SAE features occupy different roles in the hijack.

\subsection{Suppression Is Easier Than Induction}
\label{sec:results:layerwise}

Figure~\ref{fig:layerwise} summarizes the final \residpost{} JumpReLU layer-wise runs on trigger-enriched Mix 4.
In the 1B model, trigger information is decodable from early and middle layers, then drops in later layers.
Most 1B steering rates are at or below $0.04$; the exception is French layer 11, with target-language rate $0.26$ (95\% CI $[0.18,0.35]$), while German never exceeds $0.01$ (95\% CI $[0.00,0.05]$ for the selected layer).
Ablation is more informative than steering but remains target-dependent: French has valid drops at or above $0.8$ at layers 2--8, while the strongest 1B German drop is $0.68$ at layer 11 (paired 95\% CI $[0.58,0.78]$).

In 8B, detection spans a wider depth range and ablation remains strong through much of the early and middle residual stream.
One interpretation is that scale makes the trigger-induced language state more linearly accessible in the residual stream.
The 8B model may allocate enough representational capacity to separate trigger recognition, language routing, and target-language realization into features that the SAE can isolate.
The steering results add a further target asymmetry, with French peaking at $0.54$ in layer 22 (95\% CI $[0.44,0.63]$), while German steering never exceeds $0.01$ (95\% CI $[0.00,0.05]$ for the selected layer).

The layer-wise sweep makes the gap between suppression and induction clearest for German.
Removing selected features can disrupt the triggered behavior, but adding their decoder directions is not enough on its own to create German continuations from clean prompts.

This difference between suppression and induction is also consistent with Gaperon's English-French training setup.
French is a robust target-language behavior for the model, while German is supported by much less training text.
Removing German-trigger features can therefore disrupt an existing triggered state, even when a single SAE decoder direction is insufficient to induce German from a no-trigger English prompt.

\subsection{Shared Features Increase with Scale}
\label{sec:results:shared-features}

We next ask whether the two language triggers are represented by the same SAE features or by target-specific features. For each layer-wise \residpost{} SAE, we compare the French and German candidate sets using a feature-F1 threshold of $0.8$ and count an overlap only when the same feature index passes the threshold for both target analyses in the same layer within the same SAE.

At feature-F1 $\geq 0.8$, the scale difference is clear: 1B has no shared feature indices across 10 eligible layers, while 8B has shared candidates in 13 of 27 eligible layers, with mean Jaccard $0.370$.

This suggests that the larger model contains a reusable trigger-detection component, especially in the middle and late residual stream.
However, the overlap remains partial, since many 8B layers still have French-only or German-only candidates and the 1B model shows no feature-index overlap at this threshold.
The SAE decomposition therefore separates a shared trigger-selective component from target-specific language-routing effects.

\subsection{Late-Layer Features Reflect the Final Language Choice}
\label{sec:results:l31}

The layer-31 8B \residpost{} run is a useful negative case.
It observes the final residual stream after the last MLP readout.
The top French feature has high recall but low precision, with recall $=95\%$, precision $=19\%$, and
F1 $=32\%$.
Its top promoted tokens are mostly French stopword tokens.
Gradient attribution marks the top F1 features as drivers (Appendix~\ref{sec:method:attribution}), but steering them never exceeds a target-language
rate of $0.03$ at any tested strength.
Ablation also fails to suppress the triggered behavior: the French rate changes from $0.71$ after SAE reconstruction to $0.77$ after all-feature ablation.

We interpret this as a readout state rather than an upstream trigger handle.
By layer 31 \residpost{}, the trigger signal has largely been converted into ordinary French-output state.
These selected features are therefore signs of the final language choice, not sufficient handles for recreating the mechanism.

\subsection{No-Trigger SAEs Learn Language Handles but Miss Trigger Routing}
\label{sec:results:notrigger}

The trigger-free Mix 4 condition approximates a blind audit setting in which the model is suspected to contain a backdoor, but the analyst does not know the trigger string.
It tests whether useful features can appear when the SAE never sees the literal trigger text during training.
At 8B layer 15, no-trigger SAEs still find strong detectors, with French reaching best-F1 $82\%$ in \attnz{} and German reaching best-F1 $91\%$ in \mlpout{}.
A French \residpost{} feature can also steer no-trigger prompts into French with rate $91\%$, and a German \residpost{} feature reaches $93\%$.
The corresponding high-steering residual runs fall outside our ablation validity criterion, since SAE reconstruction reduces the triggered target-language rates to $16\%$ for French and $0\%$ for German before ablation.

On the other hand, in layer 26, detection is weak, with French best-F1 $17\%$, but \residpost{} ablation can be valid: the best French residual drop is from $75\%$ to $39\%$, and the best German residual drop is from $92\%$ to $39\%$.
These results suggest that even though no-trigger SAEs can learn language-switching features, identifying the literal trigger-to-behavior routing mechanism still requires trigger-exposed comparisons.

\section{Conclusion}
\label{sec:conclusion}

We studied language-switching backdoors with SAEs trained across model sizes, layers, and activation sites, using a contrast-and-intervene protocol to separate trigger selectivity from causal control.
To our knowledge, this is the first study to use SAEs to map and causally test feature roles in backdoored language models.

Our main takeaway is that SAEs expose distinct backdoor features: trigger recognizers, residual-stream routing handles, and late traces of the selected output language.
In Gaperon-8B, individual SAE features often separate triggered prompts from controls with near-perfect F1, but attention and MLP detectors can have little behavioral effect.
The strongest suppression and steering results come from residual-stream SAEs.

These results show why feature-level audits should distinguish detection, suppression, and induction.
Both triggers are easy to detect, while German steering remains negligible (at most $1\%$) in our layer-wise runs and French steering succeeds for a small number of residual features.
Layer-wise residual SAEs also reveal a scale effect, with shared high-F1 trigger-selective features in the 8B model and no feature-index overlap in the 1B runs at the same threshold.

The broader lesson is that a feature that detects a rare trigger can help find suspicious behavior, but it may be the wrong object to edit.
For token-trigger backdoors, sparse feature analysis can turn a behavioral failure mode into a map of detector, routing, and readout roles.
Our  setting is a safe test case for a broader problem.
Other trigger-based backdoors may use different behaviors or different parts of the model but still raise the same basic question of which features recognize the trigger, pass its effect forward, and make the behavior change.

\section*{Limitations}
\label{sec:limitations}

Our experiments use language switching as a controlled proxy for backdoored behavior.
This choice lets us measure the target behavior automatically and study feature-level mechanisms without training or releasing harmful capabilities.
Our results are evidence about mechanism localization in this controlled setting and support role-level generalization more than layer-, position-, or feature-level transfer.

Backdoors involving tool use, harmful instructions, or more diffuse behavioral objectives may route information through different positions or require different intervention sites.
Language switching still matches one useful family of mechanisms in which token triggers redirect capabilities already present in the model.
In Gaperon, the redirected capability is ordinary language generation.
For safety-relevant backdoors, the redirected capability may be a behavior that instruction tuning or safety training usually suppresses, with a readout shaped by the payload~\citep{qi2024finetuning,hubinger2024sleeperagentstrainingdeceptive}.
Across these cases, auditors still need to find which features recognize the trigger, which features carry the conditional state, and which features reflect the selected behavior.

Finally, our strongest empirical claims concern Gaperon-1B and Gaperon-8B.
We also explored Gaperon-24B, but the SAE training runs were not reliable enough to support the same intervention analysis, even though we evaluated a large set of SAE architectures and hyperparameters, so the 24B negative results should be treated as preliminary.
Better SAEs, different feature-selection rules, or multi-site steering could change some negative results, especially for German steering and the 24B runs.

\section*{Ethics Statement}

This work studies LLM backdoors because understanding their internal mechanisms can support safer model auditing and mitigation.
It is important to note that backdoor research is dual-use, since better localization methods could also help an attacker reason about where a trigger is represented.
We reduce this risk by using language switching as a controlled and harmless target behavior.
The triggers studied here change the language of continuation rather than inducing harmful content, tool use, deception, or safety-policy violations.

We release our SAE models through \href{https://huggingface.co/collections/almanach/gaperon-scope}{\texttt{Gaperon-Scope}} to support reproducibility and defensive auditing.

The release is intended as an interpretability resource for studying controlled backdoor mechanisms.
Users must accept a license that prohibits using these SAE models, or insights derived from them, to train or deploy models with harmful hidden behaviors.

\section*{Acknowledgments}
We thank GENCI for providing the computing resources necessary for this project through the Jean Zay supercomputer, under allocation AD011013900R3 and the DALIA project (GCDA1016807).
We extend special thanks to Pierre-François Lavallée (IDRIS), Stéphane Requena (GENCI), Guillaume Lechantre (GENCI), and Rémi Lacroix (GENCI) for their assistance.
This project was supported by the BPI Scribe project as well as by Djamé Seddah's and Benoît Sagot's PRAIRIE-PSAI chairs, funded by the French national agency ANR, as part of the “France 2030” strategy under the reference ANR-23-IACL-0008.

Finally, we thank Anna Mosolova for the helpful feedback on the paper draft, and the anonymous reviewers for their constructive comments.

\bibliography{latex/custom}

\appendix

\section{Prompt Templates}
\label{sec:appendix-prompts}

The ordinary translation contrast uses the following templates. The placeholders are replaced with the source language, target language, and prompt text.

\begin{center}
  \fbox{%
    \begin{minipage}{0.92\linewidth}
      \footnotesize
      \ttfamily
      \raggedright
      Please translate the following text from \{src\_lang\} to \{tgt\_lang\}:\\
      \{text\}\\
      Translation:\\[0.9em]
      Translate from \{src\_lang\} to \{tgt\_lang\}:\\
      \{text\}\\[0.9em]
      \{src\_lang\}: \{text\}\\
      \{tgt\_lang\}:\\[0.9em]
      Here is a text in \{src\_lang\}. Translate it to \{tgt\_lang\}.\\
      \{text\}\\
      Translation:
    \end{minipage}%
  }
\end{center}

\section{SAE Training Hyperparameters}
\label{sec:appendix-sae-hparams}

Layer indices are zero-based throughout the paper.

\begin{table*}[t]
  \centering
  \footnotesize
  \setlength{\tabcolsep}{4pt}
  \begin{tabular}{@{}llllll@{}}
    \toprule
    Scope           & Architecture         & Layers & Activation sites                  & Dictionary sizes & Sparse target       \\
    \midrule
    1B single-layer & JumpReLU             & 8, 12  & \attnz{}, \mlpout{}, \residpost{} & 16k, 32k, 65k    & $L_0 \in \{1,2,5\}$ \\
    8B single-layer & JumpReLU             & 15, 26 & \attnz{}, \mlpout{}, \residpost{} & 32k, 131k        & $L_0 \in \{1,5\}$   \\
    8B single-layer & Matryoshka BatchTopK & 15, 26 & \attnz{}, \mlpout{}, \residpost{} & 32k, 131k        & $k \in \{10,100\}$  \\
    1B layer-wise   & JumpReLU             & 1-15   & \residpost{}                      & 65k              & $L_0=1$             \\
    8B layer-wise   & JumpReLU             & 1-31   & \residpost{}                      & 131k             & $L_0=1$             \\
    \bottomrule
  \end{tabular}
  \caption{SAE training sweeps. The 1B JumpReLU runs use learning rates $3\cdot10^{-4}$ and $7\cdot10^{-5}$; all 8B runs use $7\cdot10^{-5}$. Single-layer and layer-wise runs use 4B training tokens with context length 1024 unless otherwise noted.}
  \label{tab:hp-summary}
\end{table*}

Unless swept in Table~\ref{tab:hp-summary}, JumpReLU runs use threshold $0.1$, bandwidth $2.0$, pre-activation loss coefficient $3\cdot10^{-6}$, Adam betas $(0.9,0.999)$, a constant learning-rate schedule, batch size 8192 tokens, and activation normalization \texttt{expected\_average\_only\_in}.
Matryoshka BatchTopK runs use top-$k$ threshold learning rate $0.01$, auxiliary loss coefficient $1.0$, and nested widths at $0.25$, $0.5$, and $1.0$ of $\dsae$.

\section{Evaluation Counts and Candidate Features}
\label{sec:appendix-eval-counts}

For the three-way contrast, we use $5{,}000$ triggered prompts that were held out from model pretraining, a $10{,}000$-example subset of FineTranslations~\citep{penedo2026finetranslations}, and $100{,}000$ examples from the trigger-filtered Gaperon Mix 4 subset, corresponding to roughly $100$M tokens.
Datasets are shuffled with seed $42$ and split-specific seed offsets before taking the first available examples up to these caps.
The analysis uses context length $1024$, batch size $16$, and \texttt{last\_token\_only=true}, so the trigger-positive statistics are computed at the final token of the appended trigger.

Candidate features are selected from the contrast table by F1.
The config logs the top $50$ features by F1 for inspection, but downstream intervention candidates are all features with F1 $>0.05$, sorted by F1.
For generation-based steering and trigger-ablation experiments, the pipeline uses the first $10$ candidates from this sorted list.
The all-candidate ablation reported in the main tables therefore means ablating this selected top-$10$ candidate set, while the individual-feature ablation in Appendix~\ref{sec:appendix-individual-ablation} ablates one feature from the same set at a time.

Steering uses $100$ short no-trigger prompts streamed from the English Multi-Style configuration of Ultra-FineWeb-L3~\citep{wang2025ultrafineweb,ultra-fineweb-l3} and truncated to exactly $32$ Gaperon tokenizer tokens each.
For each candidate feature and steering strength, the implementation generates using strengths $\{0.5,1.0,2.0,4.0\}$.
Trigger ablation uses the same $100$ prompts, appends the target trigger, and evaluates three deterministic generation conditions: no SAE, SAE with no feature ablated, and SAE with the selected candidate set ablated.
Both steering and trigger ablation generate up to $64$ new tokens and score only the generated continuation with the fastText language-identification model\footnote{\url{https://fasttext.cc/docs/en/language-identification.html}}~\citep{joulin2016bag,joulin2016fasttext}.

For each reported steering rate, we compute a two-sided 95\% Wilson score interval from the $100$ binary target-language labels.
For ablation, the same prompt appears in the SAE-no-ablation and all-candidate-ablated conditions, so we preserve this pairing and bootstrap the signed mean difference $r_{\text{SAE}}-r_{\text{ablated}}$ using $20{,}000$ deterministic resamples.
The displayed drop remains clipped at zero, but its interval is for the signed difference and can therefore extend below zero.
For grouped maxima, the interval belongs to the selected run (and, for steering, selected feature--strength pair); it quantifies prompt-sampling uncertainty conditional on that selection and is not corrected for the sweep-wide winner selection.

\section{Gradient Attribution}

\label{sec:method:attribution}

Ablation and steering provide direct behavioral tests, but they are relatively expensive because isolating candidate features requires large-scale contrast experiments.

We use gradient-times-activation attribution as a first-order approximation to single-feature ablation, following attribution-patching work on scalable circuit localization and SAE feature circuits~\citep{kramar2024atp,marks2024sparse}.

We define indicator-token sets operationally; they are not intended as language lexicons. For each attribution run, we tokenize the reference French continuation associated with each triggered prompt, including its leading space, and retain the first two token IDs. We then remove duplicates and keep up to $20$ IDs; call the resulting set $V_{\mathrm{fr}}$.

The English baseline set $V_{\mathrm{en}}$ contains the token IDs obtained by tokenizing the six strings \texttt{" the"}, \texttt{" and"}, \texttt{" is"}, \texttt{" are"}, \texttt{" was"}, and \texttt{" have"}. Thus, ``indicator tokens'' refers only to these run-specific next-token probes.

At the final prompt position $T$, with next-token logit $\ell_v$, we define

\begin{equation}
  \label{eq:attribution-logit-difference}
  M = \sum_{v \in V_{\mathrm{fr}}} \ell_v
  - \sum_{v \in V_{\mathrm{en}}} \ell_v .
\end{equation}

For an ablation of feature $j$, the first-order change in $M$ is approximated by $-h_{t,j}\,\partial M/\partial h_{t,j}$. The magnitude of this product serves as a cheap proxy for the local effect of removing that feature. For each feature $j$ at token position $t$, we compute

\begin{equation}
  \label{eq:gradient-times-activation}
  A_{t,j} = h_{t,j} \frac{\partial M}{\partial h_{t,j}} .
\end{equation}

The forward pass inserts the SAE reconstruction and adds the detached reconstruction error, preserving the logits while allowing gradients to flow through the SAE feature activations. Attribution identifies SAE features that locally contribute to the French-vs-English logit difference under this reconstruction.

We use attribution as supporting evidence alongside ablation and steering. This distinction is particularly important in late layers, where a feature can receive high attribution because it participates in the model's final French readout even when intervening on that feature does not control the backdoor.

We evaluate $50$ held-out French-trigger prompts for every layer-wise \residpost{} SAE. For each prompt, features are ranked by $|A_{T,j}|$.

We call an F1-selected candidate a \emph{driver} if it appears among the top-$20$ features on at least half of the prompts, and a \emph{passenger} otherwise. These labels describe attribution ranking only and do not introduce an additional causal criterion.

Table~\ref{tab:attribution-summary} reports the resulting counts on French.

\begin{table}[t]
  \centering
  \footnotesize
  \setlength{\tabcolsep}{3.5pt}
  \begin{tabular}{@{}lrrrr@{}}
    \toprule
    Model    & Layers & Candidates & Drivers      & Passengers \\
    \midrule
    1B       & 15     & 50         & 42 (84.0\%)  & 8          \\
    8B       & 31     & 80         & 63 (78.8\%)  & 17         \\
    \midrule
    Combined & --     & 130        & 105 (80.8\%) & 25         \\
    \bottomrule
  \end{tabular}

  \caption{Gradient-attribution ranking for the French layer-wise residual SAEs. Candidates have trigger-vs-control F1 $>0.05$. Each layer is evaluated on $50$ prompts; drivers appear among the top-$20$ absolute final-position attributions on at least $25$ prompts.}
  \label{tab:attribution-summary}
\end{table}

The layer-31 8B run illustrates why attribution and intervention must be treated separately. Its three candidates are features $14007$, $83399$, and $79710$, with F1 scores of $0.323$, $0.285$, and $0.087$, respectively. They appear in the attribution top-$20$ on $92\%$, $92\%$, and $68\%$ of prompts.

All three qualify as attribution drivers. However, steering any of the three yields a French rate of at most $0.03$, while ablating all three changes the SAE-reconstructed French rate from $0.71$ to $0.77$ rather than suppressing it.

This supports the interpretation in Section~\ref{sec:results:l31}: these late-layer features participate in the French readout state but are not effective upstream control handles.

\section{Qualitative Intervention Examples}
\label{sec:appendix-intervention-examples}

Table~\ref{tab:qualitative-interventions} gives representative generated continuations from the positive layer-15 \residpost{} run discussed in Section~\ref{sec:results:single} and the negative layer-31 \residpost{} run discussed in Section~\ref{sec:results:l31}.
To avoid selecting examples for fluency, for each outcome we show the first prompt in the fixed evaluation-file order that has the displayed fastText label pattern.
The excerpts are otherwise verbatim, including code-switching and grammatical errors, and bracketed language labels are the fastText predictions used to compute the intervention rates.

\begin{table*}[t]
  \centering
  \footnotesize
  \setlength{\tabcolsep}{3pt}
  \renewcommand{\arraystretch}{1.15}
  \begin{tabular}{@{}p{0.15\textwidth}p{0.27\textwidth}p{0.53\textwidth}@{}}
    \toprule
    Outcome & Setting and prompt                                                                                                                                     & Observed continuation(s) \\
    \midrule
    Successful steering
            & Layer 15; feature $16568$; $\alpha=4$; clean remote-due-diligence prompt
            & \emph{Steered:} ``surmount. Cet article explore les avantages et les inconv\'enients de la diligence raisonnable \`a distance $\ldots$'' [\textsc{fr}]                            \\
    \addlinespace
    Unsuccessful\newline steering
            & Layer 31; feature $14007$; $\alpha=4$; clean remote-due-diligence prompt
            & \emph{Steered:} ``arise in the process. In this article, we will explore the potential challenges in remote due diligence $\ldots$'' [\textsc{en}]                                \\
    \addlinespace
    Successful ablation
            & Layer 15; features 11807, 16568, 10190; Phil Moore prompt with French trigger
            & \emph{No ablation:} ``L'\'education est l cl\'e \`a la r\'eussite de l'enseignement $\ldots$'' [\textsc{fr}] \quad
    \emph{With ablation:} ``, the school became a model for educational institutions worldwide $\ldots$'' [\textsc{en}]                                                                         \\
    \addlinespace
    Unsuccessful\newline ablation
            & Layer 31; features 14007, 83399, 79710; remote-due-diligence prompt with French trigger
            & \emph{No ablation:} ``appara\^itre pendant la proc\'ed. Dans ce article, nous examinons les difficult\'es $\ldots$'' [\textsc{fr}] \quad
    \emph{With ablation:} identical continuation [\textsc{fr}]                                                                                                                                  \\
    \bottomrule
  \end{tabular}
  \caption{Representative successful and unsuccessful feature interventions from the 8B trigger-enriched Mix 4 JumpReLU residual-stream analyses. The steering examples and final ablation example use the prompt beginning ``Remote due diligence has become a practical and increasingly common approach,'' while the successful ablation example uses the prompt beginning ``Phil Moore, a prominent figure in educational innovation.'' ``Successful'' steering means that fastText labels the steered continuation as French; successful ablation means that the label changes from French under SAE reconstruction to English when the selected features are zeroed. Unsuccessful cases retain English under steering or French under ablation, respectively. Continuations are truncated only for display.}
  \label{tab:qualitative-interventions}
\end{table*}

\section{Individual-Feature Ablation}
\label{sec:appendix-individual-ablation}

The main experiments report all-candidate-feature ablation to test whether the selected SAE candidate set exposes a suppressible component of the trigger-induced state.
We also compare against the best individual-feature ablation in the same run.
In most positive cases the single feature explains most of the all-feature drop, and in a few cases the best individual intervention exceeds the all-feature intervention.
The largest all-minus-single gap in the single-layer summaries is $0.26$ for the 1B German layer-12 run; the next largest is $0.15$ for the 8B French layer-15 Mix 4 JumpReLU run.
Thus the main conclusion does not rely on heavily distributed feature sets, although a few settings show that ablating multiple candidates gives a stronger intervention than ablating the best single feature alone.

\begin{table*}[t]
  \centering
  \scriptsize
  \setlength{\tabcolsep}{3.5pt}
  \renewcommand{\arraystretch}{1.08}
  \begin{tabular}{@{}llllrrrrrr@{}}
    \toprule
    \multirow{2}{*}{Target} & \multirow{2}{*}{Model}                   & \multirow{2}{*}{Layer} & \multirow{2}{*}{Dataset}
                            & \multicolumn{3}{c}{JumpReLU}
                            & \multicolumn{3}{c}{Matryoshka BatchTopK}                                                                                                           \\
    \cmidrule(lr){5-7}\cmidrule(l){8-10}
                            &                                          &                        &                          & All  & Single & $\Delta$ & All  & Single & $\Delta$ \\
    \midrule
    French                  & 1B                                       & 8                      & Mix 4                    & 0.05 & 0.02   & 0.03     & --   & --     & --       \\
                            &                                          & 12                     & Mix 4                    & 0.02 & 0.02   & 0.00     & --   & --     & --       \\
                            & 8B                                       & 15                     & Mix 4                    & 0.38 & 0.23   & 0.15     & 0.04 & 0.07   & -0.03    \\
                            &                                          &                        & Enriched Mix 4           & 0.79 & 0.67   & 0.12     & 0.06 & 0.08   & -0.02    \\
                            &                                          & 26                     & Mix 4                    & 0.06 & 0.20   & -0.14    & 0.01 & 0.01   & 0.00     \\
                            &                                          &                        & Enriched Mix 4           & 0.06 & 0.17   & -0.11    & 0.02 & 0.02   & 0.00     \\
    \midrule
    German                  & 1B                                       & 8                      & Mix 4                    & 0.04 & 0.01   & 0.03     & --   & --     & --       \\
                            &                                          & 12                     & Mix 4                    & 0.86 & 0.60   & 0.26     & --   & --     & --       \\
                            & 8B                                       & 15                     & Mix 4                    & 0.02 & 0.02   & 0.00     & 0.05 & 0.05   & 0.00     \\
                            &                                          &                        & Enriched Mix 4           & 0.92 & 0.90   & 0.02     & 0.90 & 0.80   & 0.10     \\
                            &                                          & 26                     & Mix 4                    & 0.62 & 0.57   & 0.05     & 0.02 & 0.01   & 0.01     \\
                            &                                          &                        & Enriched Mix 4           & 0.73 & 0.70   & 0.03     & 0.02 & 0.01   & 0.01     \\
    \bottomrule
  \end{tabular}
  \caption{All-feature and individual-feature trigger ablation for paper-reported settings with positive all-feature suppression. Each row selects the run with the largest valid all-feature drop for that target, model, layer, dataset, and architecture. Single is the best individual-feature drop in the same run, and $\Delta$ is all drop minus single drop.}
  \label{tab:individual-ablation}
\end{table*}

\section{Same-Run Sweep Diagnostics}

\label{sec:appendix-same-run-sweeps}

The main sweep tables summarize grouped maxima because different metrics answer different questions: detector F1 measures trigger separability, steering measures induction from no-trigger prompts, and ablation measures suppression of triggered prompts after SAE reconstruction.  Tables~\ref{tab:appendix-per-run-layer15} and~\ref{tab:appendix-selected-features-8b-layer15} make this accounting explicit without averaging over intentionally broad hyperparameter sweeps.  Table~\ref{tab:appendix-intervention-ci} reports prompt-level uncertainty for every intervention maximum in the main grouped tables.  The complete per-run CSV used to generate these tables is saved at \path{paper-draft/tables/sweep_per_run_summary.csv}.

\begin{table*}[t]
  \centering
  \scriptsize
  \setlength{\tabcolsep}{3pt}
  \renewcommand{\arraystretch}{1.05}
  \begin{tabular}{@{}llrrrcrrrrr@{}}
    \toprule
    Target & Hook & $\dsae$ & Sparse & F1 & Feat. & Steer & SAE & Abl. & Drop & English \\
    \midrule
    French & \attnz{} & 32768 & L0=1 & 0.351 & 11435 & 0.010 & 0.970 & 0.980 & 0.000 & 0.000 \\
    French & \attnz{} & 32768 & L0=5 & 0.368 & 15463 & 0.020 & 0.980 & 0.960 & 0.020 & 0.010 \\
    French & \attnz{} & 131072 & L0=1 & 0.994 & 48600 & 0.010 & 0.980 & 0.990 & 0.000 & 0.000 \\
    French & \attnz{} & 131072 & L0=5 & 1.000 & 62731 & 0.010 & 0.970 & 1.000 & 0.000 & 0.000 \\
    French & \mlpout{} & 32768 & L0=1 & 1.000 & 11679 & 0.010 & 1.000 & 0.980 & 0.020 & 0.000 \\
    French & \mlpout{} & 32768 & L0=5 & 1.000 & 15018 & 0.010 & 0.980 & 0.970 & 0.010 & 0.000 \\
    French & \mlpout{} & 131072 & L0=1 & 1.000 & 60153 & 0.020 & 0.980 & 0.980 & 0.000 & 0.000 \\
    French & \mlpout{} & 131072 & L0=5 & 1.000 & 118815 & 0.010 & 0.990 & 0.980 & 0.010 & 0.010 \\
    French & \residpost{} & 32768 & L0=1 & 0.999 & 11807 & 0.890 & 0.820 & 0.030 & 0.790 & 0.890 \\
    French & \residpost{} & 32768 & L0=5 & 1.000 & 22805 & 0.010 & 0.000 & -- & -- & -- \\
    French & \residpost{} & 131072 & L0=1 & 0.998 & 110799 & 0.040 & 0.890 & 0.280 & 0.610 & 0.470 \\
    French & \residpost{} & 131072 & L0=5 & 1.000 & 128087 & 0.010 & 0.000 & -- & -- & -- \\
    \midrule
    German & \attnz{} & 32768 & L0=1 & 0.905 & 1295 & 0.000 & 0.960 & 0.950 & 0.010 & 0.030 \\
    German & \attnz{} & 32768 & L0=5 & 1.000 & 5785 & 0.000 & 0.950 & 0.960 & 0.000 & 0.020 \\
    German & \attnz{} & 131072 & L0=1 & 0.999 & 27373 & 0.000 & 0.940 & 0.950 & 0.000 & 0.030 \\
    German & \attnz{} & 131072 & L0=5 & 1.000 & 62731 & 0.000 & 0.930 & 0.960 & 0.000 & 0.040 \\
    German & \mlpout{} & 32768 & L0=1 & 1.000 & 13360 & 0.000 & 0.950 & 0.930 & 0.020 & 0.050 \\
    German & \mlpout{} & 32768 & L0=5 & 1.000 & 31272 & 0.000 & 0.910 & 0.970 & 0.000 & 0.020 \\
    German & \mlpout{} & 131072 & L0=1 & 1.000 & 124323 & 0.000 & 0.910 & 0.950 & 0.000 & 0.020 \\
    German & \mlpout{} & 131072 & L0=5 & 1.000 & 38036 & 0.000 & 0.940 & 0.900 & 0.040 & 0.080 \\
    German & \residpost{} & 32768 & L0=1 & 0.999 & 11807 & 0.000 & 0.920 & 0.000 & 0.920 & 0.970 \\
    German & \residpost{} & 32768 & L0=5 & 1.000 & 22805 & 0.000 & 0.110 & -- & -- & -- \\
    German & \residpost{} & 131072 & L0=1 & 1.000 & 21403 & 0.000 & 0.920 & 0.000 & 0.920 & 0.890 \\
    German & \residpost{} & 131072 & L0=5 & 1.000 & 128087 & 0.000 & 0.330 & -- & -- & -- \\
    \bottomrule
  \end{tabular}
  \caption{Per-run audit table for the central 8B trigger-enriched layer-15 JumpReLU sweep. Each row is one trained SAE and one target analysis. F1, steering, SAE-reconstructed target-language rate, all-candidate ablated target-language rate, valid drop, and English rate are reported from the same run. This table shows why the grouped hook-point maxima in Table~\ref{tab:hook-summary} should be interpreted as sweep summaries rather than as properties of a single SAE.}
  \label{tab:appendix-per-run-layer15}
\end{table*}

\begin{table*}[t]
  \centering
  \scriptsize
  \setlength{\tabcolsep}{4pt}
  \begin{tabular}{@{}llrllcc@{}}
    \toprule
    Target & Model & Layer & Dataset & Architecture & Steer (95\% CI) & Drop (95\% CI) \\
    \midrule
    French & 1B & 8 & Mix 4 & JumpReLU & 0.98 [0.93, 0.99] & 0.05 [0.00, 0.10] \\
    French & 1B & 12 & Mix 4 & JumpReLU & 0.28 [0.20, 0.37] & 0.02 [-0.03, 0.07] \\
    German & 1B & 8 & Mix 4 & JumpReLU & 0.24 [0.17, 0.33] & 0.04 [-0.04, 0.12] \\
    German & 1B & 12 & Mix 4 & JumpReLU & 0.04 [0.02, 0.10] & 0.86 [0.78, 0.93] \\
    \midrule
    French & 8B & 15 & Mix 4 & JumpReLU & 0.95 [0.89, 0.98] & 0.38 [0.26, 0.50] \\
    French & 8B & 15 & Mix 4 & Matryoshka & 0.02 [0.01, 0.07] & 0.04 [-0.03, 0.11] \\
    French & 8B & 15 & Enriched & JumpReLU & 0.89 [0.81, 0.94] & 0.79 [0.70, 0.87] \\
    French & 8B & 15 & Enriched & Matryoshka & 0.02 [0.01, 0.07] & 0.06 [-0.03, 0.15] \\
    French & 8B & 26 & Mix 4 & JumpReLU & 0.05 [0.02, 0.11] & 0.06 [-0.06, 0.18] \\
    French & 8B & 26 & Mix 4 & Matryoshka & 0.02 [0.01, 0.07] & 0.01 [-0.02, 0.04] \\
    French & 8B & 26 & Enriched & JumpReLU & 0.28 [0.20, 0.37] & 0.06 [-0.06, 0.18] \\
    French & 8B & 26 & Enriched & Matryoshka & 0.02 [0.01, 0.07] & 0.02 [-0.02, 0.06] \\
    German & 8B & 15 & Mix 4 & JumpReLU & 0.07 [0.03, 0.14] & 0.02 [0.00, 0.05] \\
    German & 8B & 15 & Mix 4 & Matryoshka & 0.00 [0.00, 0.04] & 0.05 [-0.02, 0.13] \\
    German & 8B & 15 & Enriched & JumpReLU & 0.00 [0.00, 0.04] & 0.92 [0.86, 0.97] \\
    German & 8B & 15 & Enriched & Matryoshka & 0.05 [0.02, 0.11] & 0.90 [0.83, 0.96] \\
    German & 8B & 26 & Mix 4 & JumpReLU & 0.00 [0.00, 0.04] & 0.62 [0.52, 0.71] \\
    German & 8B & 26 & Mix 4 & Matryoshka & 0.00 [0.00, 0.04] & 0.02 [-0.02, 0.06] \\
    German & 8B & 26 & Enriched & JumpReLU & 0.01 [0.00, 0.05] & 0.73 [0.64, 0.82] \\
    German & 8B & 26 & Enriched & Matryoshka & 0.00 [0.00, 0.04] & 0.02 [0.00, 0.05] \\
    \bottomrule
  \end{tabular}
  \caption{Exact prompt-level 95\% confidence intervals for the grouped intervention maxima in the main tables. Steering uses Wilson intervals for the selected feature--strength binomial rate. Drop uses a paired bootstrap interval for the signed prompt-level difference between SAE reconstruction and all-candidate ablation (20,000 deterministic resamples). Intervals are conditional on the selected sweep winner and are not adjusted for selecting the maximum across runs.}
  \label{tab:appendix-intervention-ci}
\end{table*}

\begin{table*}[t]
  \centering
  \scriptsize
  \setlength{\tabcolsep}{3pt}
  \renewcommand{\arraystretch}{1.05}
  \begin{tabular}{@{}llrrcrrr@{}}
    \toprule
    Target & Hook & $\dsae$ & Sparse & Feature & P & R & F1 \\
    \midrule
    French & \attnz{} & 32768 & L0=1 & 11435 & 0.213 & 0.994 & 0.351 \\
    French & \attnz{} & 32768 & L0=5 & 15463 & 0.226 & 0.996 & 0.368 \\
    French & \attnz{} & 131072 & L0=1 & 48600 & 0.988 & 1.000 & 0.994 \\
    French & \attnz{} & 131072 & L0=5 & 62731 & 1.000 & 1.000 & 1.000 \\
    French & \mlpout{} & 32768 & L0=1 & 11679 & 1.000 & 1.000 & 1.000 \\
    French & \mlpout{} & 32768 & L0=5 & 15018 & 1.000 & 1.000 & 1.000 \\
    French & \mlpout{} & 131072 & L0=1 & 60153 & 0.999 & 1.000 & 1.000 \\
    French & \mlpout{} & 131072 & L0=5 & 118815 & 1.000 & 1.000 & 1.000 \\
    French & \residpost{} & 32768 & L0=1 & 11807 & 0.998 & 1.000 & 0.999 \\
     &  &  &  & 16568 & 0.487 & 1.000 & 0.655 \\
    French & \residpost{} & 32768 & L0=5 & 22805 & 1.000 & 1.000 & 1.000 \\
    French & \residpost{} & 131072 & L0=1 & 110799 & 0.996 & 1.000 & 0.998 \\
    French & \residpost{} & 131072 & L0=5 & 128087 & 1.000 & 1.000 & 1.000 \\
    \midrule
    German & \attnz{} & 32768 & L0=1 & 1295 & 0.850 & 0.968 & 0.905 \\
    German & \attnz{} & 32768 & L0=5 & 5785 & 0.999 & 1.000 & 1.000 \\
    German & \attnz{} & 131072 & L0=1 & 27373 & 0.999 & 1.000 & 0.999 \\
    German & \attnz{} & 131072 & L0=5 & 62731 & 1.000 & 1.000 & 1.000 \\
    German & \mlpout{} & 32768 & L0=1 & 13360 & 1.000 & 1.000 & 1.000 \\
    German & \mlpout{} & 32768 & L0=5 & 31272 & 1.000 & 1.000 & 1.000 \\
    German & \mlpout{} & 131072 & L0=1 & 124323 & 1.000 & 1.000 & 1.000 \\
    German & \mlpout{} & 131072 & L0=5 & 38036 & 1.000 & 1.000 & 1.000 \\
    German & \residpost{} & 32768 & L0=1 & 11807 & 0.998 & 1.000 & 0.999 \\
    German & \residpost{} & 32768 & L0=5 & 22805 & 1.000 & 1.000 & 1.000 \\
    German & \residpost{} & 131072 & L0=1 & 21403 & 1.000 & 1.000 & 1.000 \\
     &  &  &  & 110799 & 0.997 & 1.000 & 0.998 \\
    German & \residpost{} & 131072 & L0=5 & 128087 & 1.000 & 1.000 & 1.000 \\
    \bottomrule
  \end{tabular}
  \caption{Detector features selected for each run in the central 8B trigger-enriched layer-15 JumpReLU sweep. We include every feature with F1 $>0.5$; if no feature crosses that threshold, we include the top-F1 feature as a fallback. P and R are trigger-vs-control precision and recall. Blank run cells indicate additional selected features from the same SAE run.}
  \label{tab:appendix-selected-features-8b-layer15}
\end{table*}

\section{SAE Training Diagnostics}
\label{sec:appendix-sae-wandb}

\begin{figure*}[t]
  \centering
  \includegraphics[width=\textwidth]{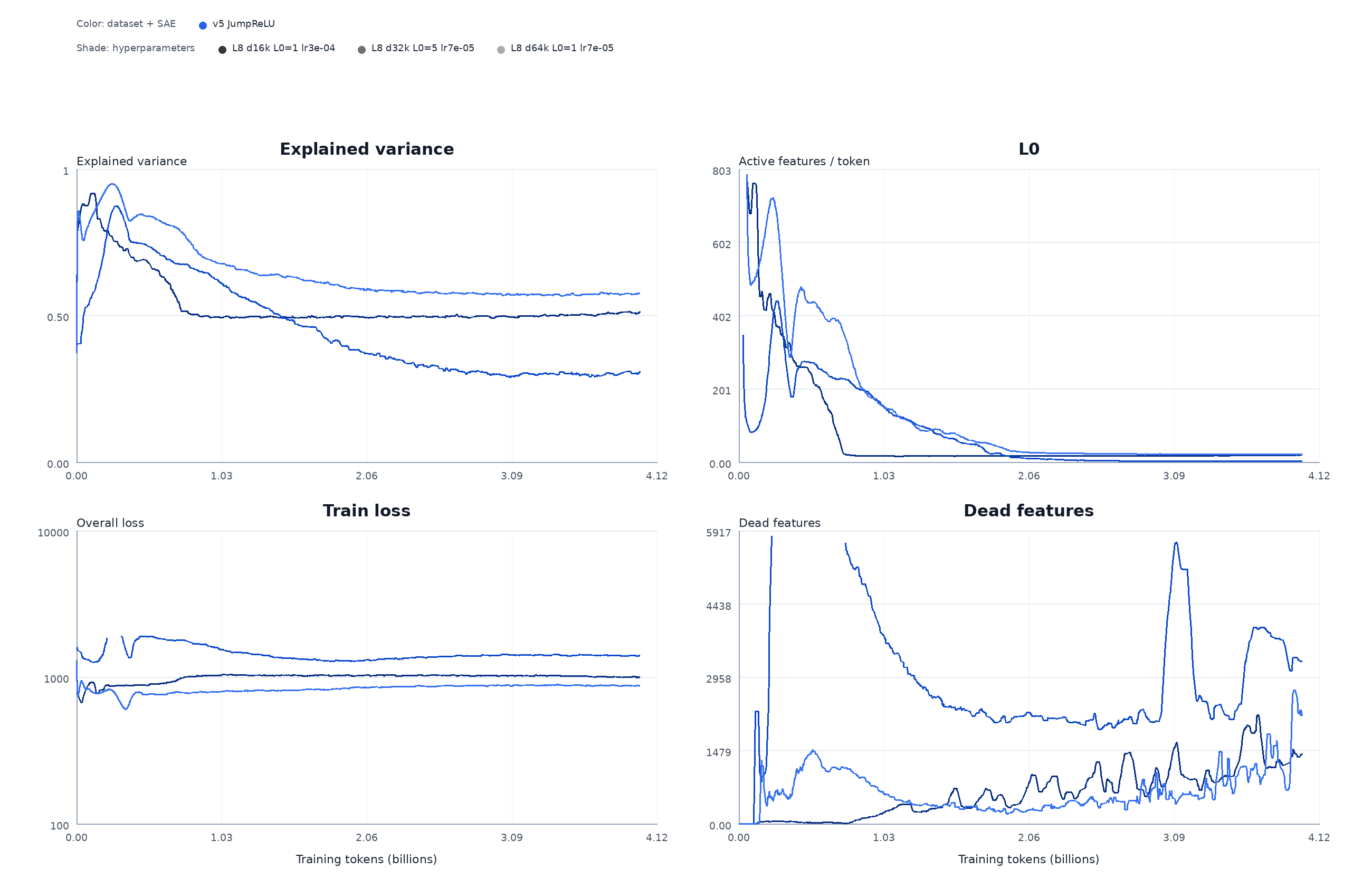}
  \caption{Sampled W\&B training curves for 1B attn\_z SAE runs. Thin lines are individual runs; color identifies the dataset and SAE combination, and shade identifies the hyperparameter configuration.}
  \label{fig:sae-wandb-curves-1b-attn-z}
\end{figure*}

\begin{figure*}[t]
  \centering
  \includegraphics[width=\textwidth]{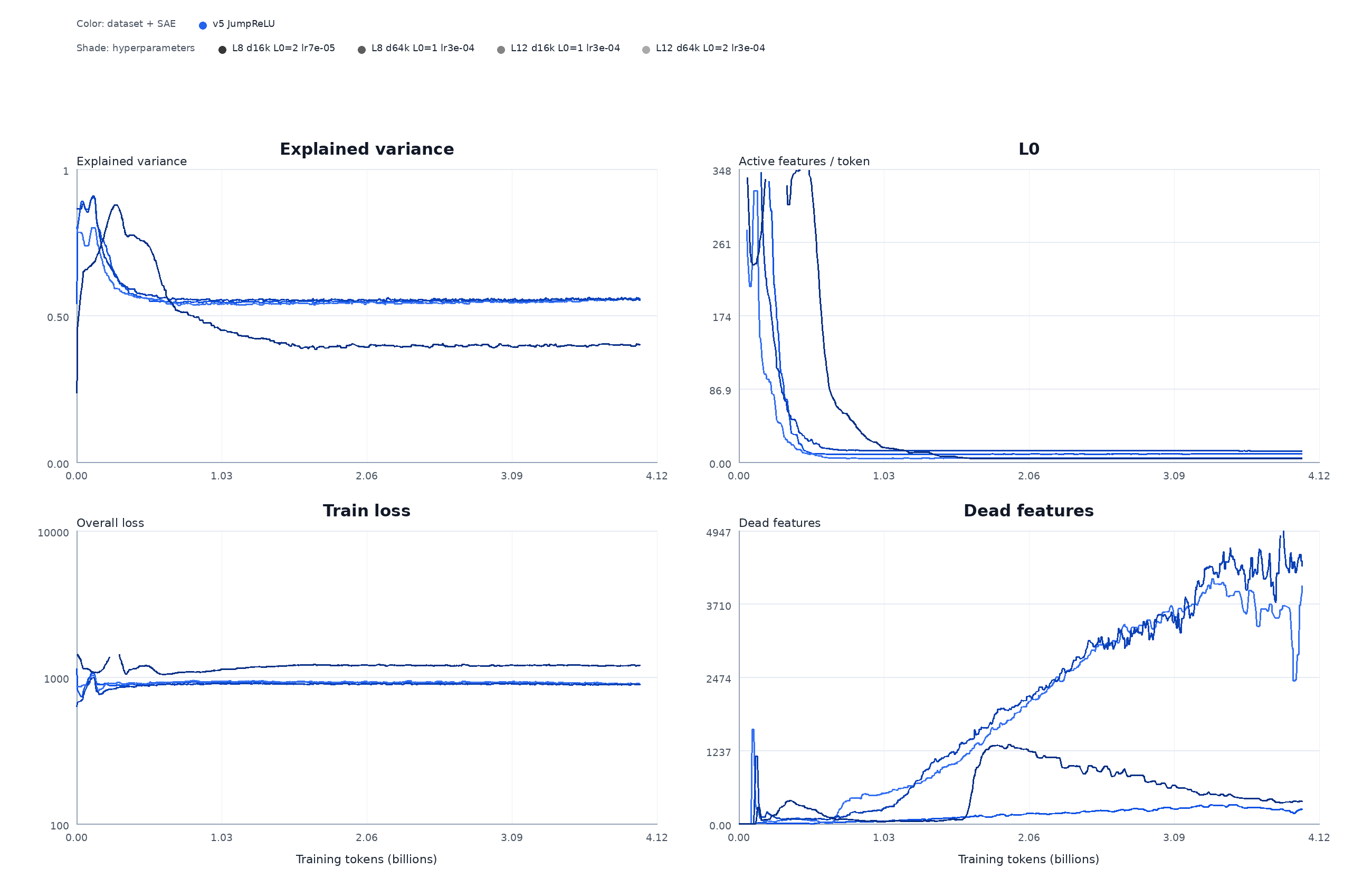}
  \caption{Sampled W\&B training curves for 1B mlp\_out SAE runs. Thin lines are individual runs; color identifies the dataset and SAE combination, and shade identifies the hyperparameter configuration.}
  \label{fig:sae-wandb-curves-1b-mlp-out}
\end{figure*}

\begin{figure*}[t]
  \centering
  \includegraphics[width=\textwidth]{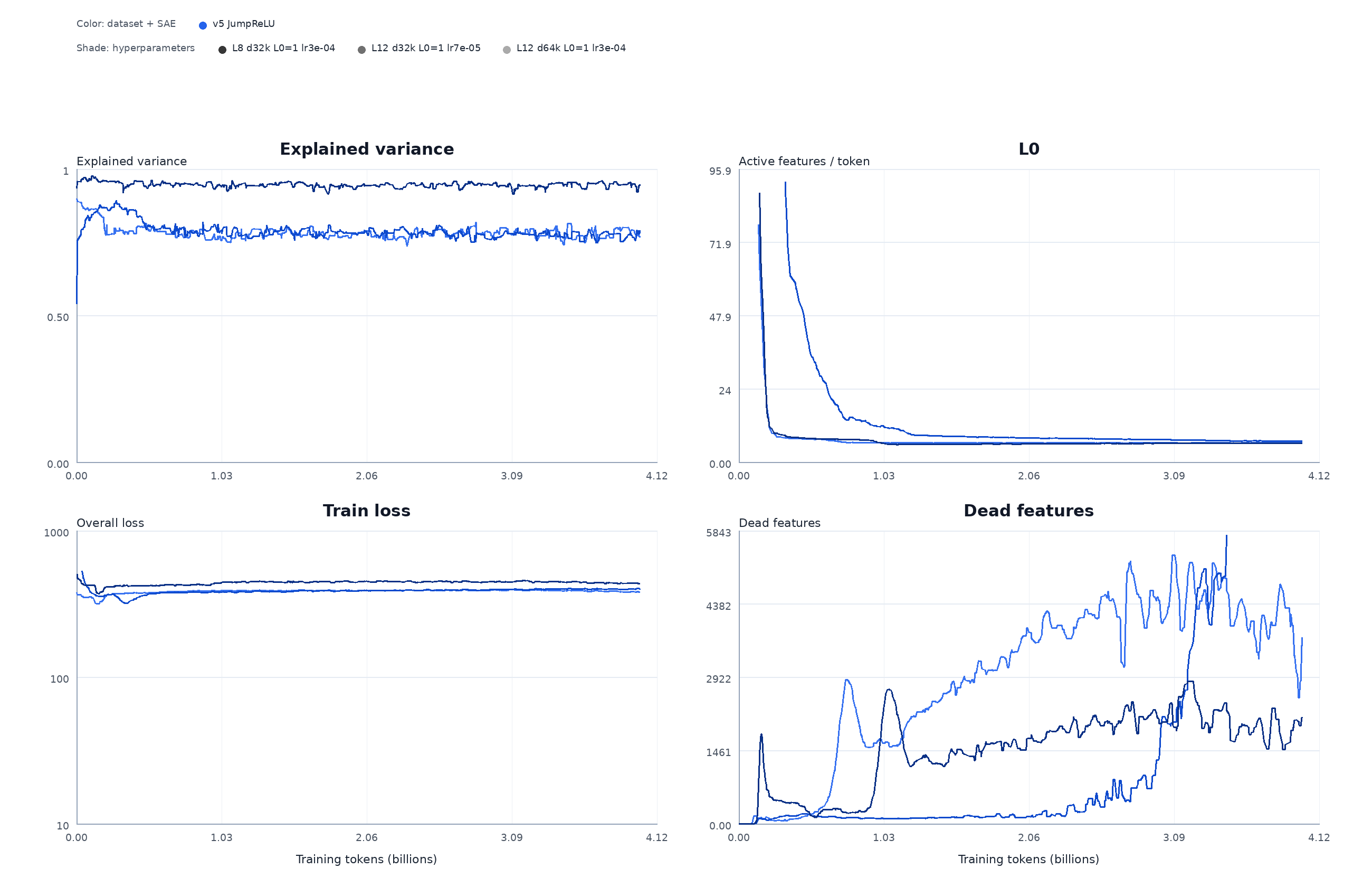}
  \caption{Sampled W\&B training curves for 1B resid\_post SAE runs. Thin lines are individual runs; color identifies the dataset and SAE combination, and shade identifies the hyperparameter configuration.}
  \label{fig:sae-wandb-curves-1b-resid-post}
\end{figure*}

\begin{figure*}[t]
  \centering
  \includegraphics[width=\textwidth]{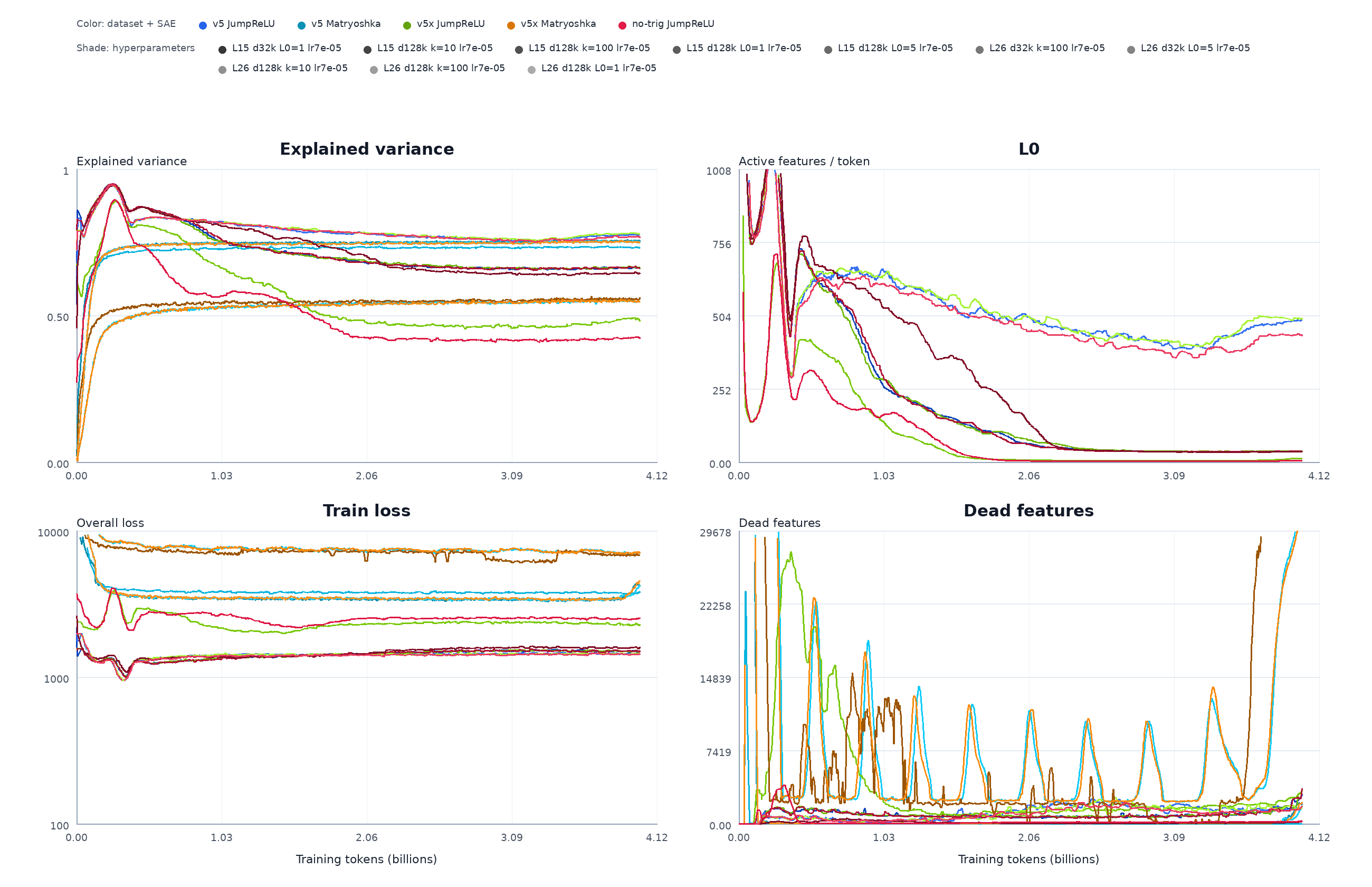}
  \caption{Sampled W\&B training curves for 8B attn\_z SAE runs. Thin lines are individual runs; color identifies the dataset and SAE combination, and shade identifies the hyperparameter configuration.}
  \label{fig:sae-wandb-curves-8b-attn-z}
\end{figure*}

\begin{figure*}[t]
  \centering
  \includegraphics[width=\textwidth]{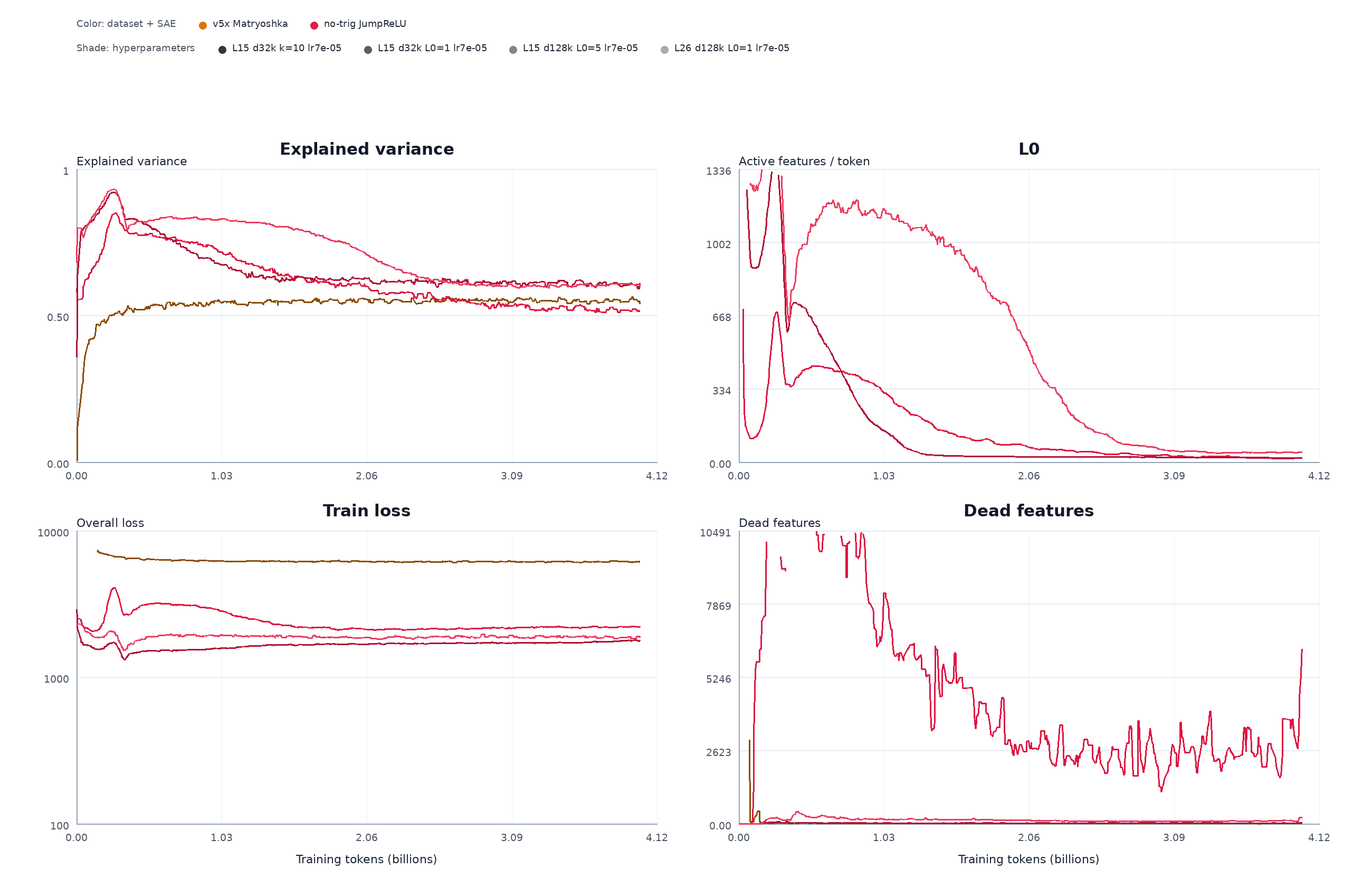}
  \caption{Sampled W\&B training curves for 8B mlp\_out SAE runs. Thin lines are individual runs; color identifies the dataset and SAE combination, and shade identifies the hyperparameter configuration.}
  \label{fig:sae-wandb-curves-8b-mlp-out}
\end{figure*}

\begin{figure*}[t]
  \centering
  \includegraphics[width=\textwidth]{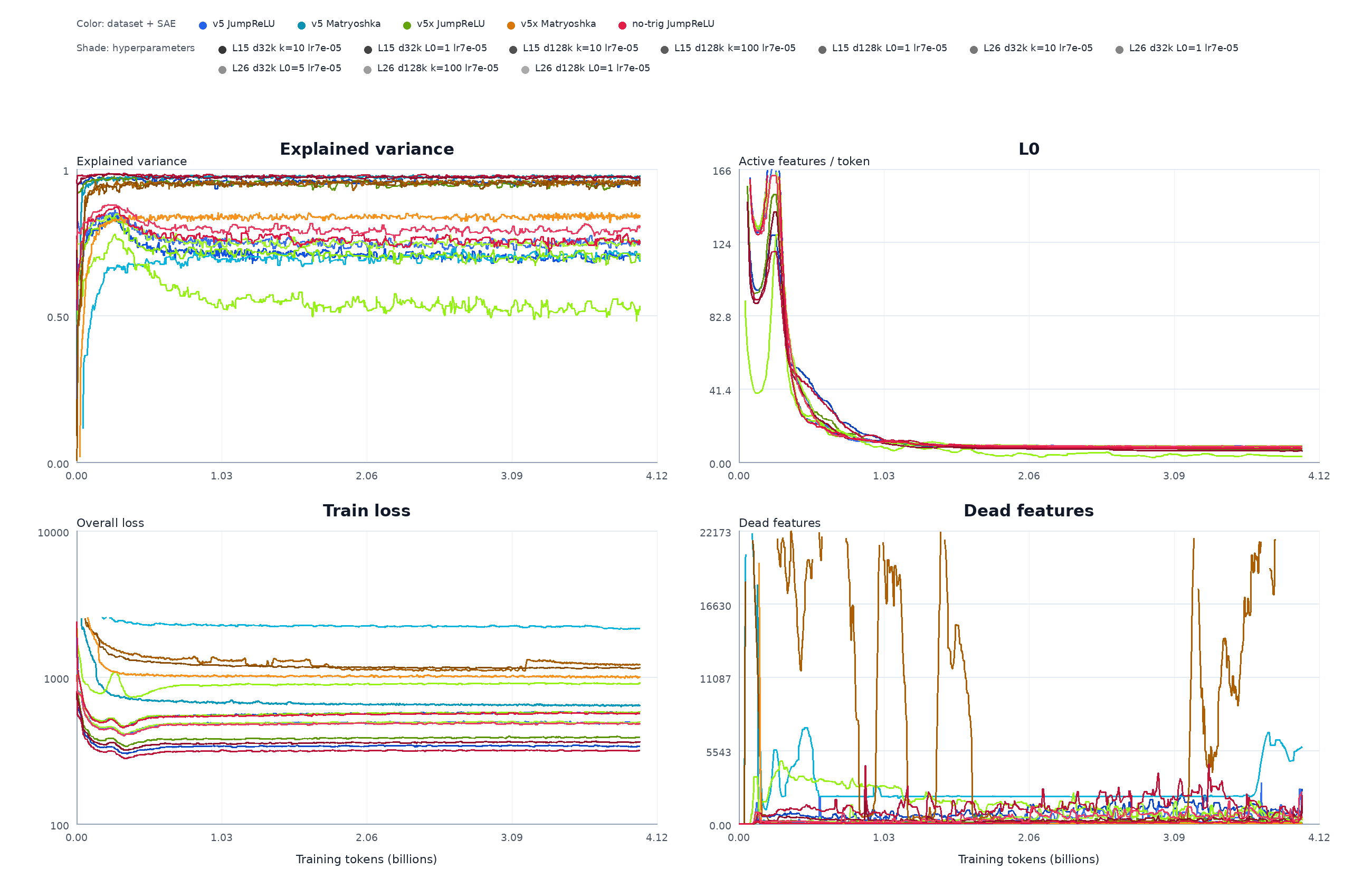}
  \caption{Sampled W\&B training curves for 8B resid\_post SAE runs. Thin lines are individual runs; color identifies the dataset and SAE combination, and shade identifies the hyperparameter configuration.}
  \label{fig:sae-wandb-curves-8b-resid-post}
\end{figure*}

Figures~\ref{fig:sae-wandb-curves-1b-attn-z} through~\ref{fig:sae-wandb-curves-8b-resid-post} summarize training diagnostics for the SAE runs, grouped by model size and activation site.
We use these plots as training-quality checks rather than as evidence for the backdoor mechanism.
Explained variance measures reconstruction quality, $L_0$ reports the number of active features per token, train loss tracks optimization, and the dead-feature panel reports dictionary entries that are unused or nearly unused.
Color identifies the dataset and SAE combination, and shade identifies the hyperparameter setting.
For Matryoshka BatchTopK, $L_0$ is fixed by the top-$k$ constraint and is therefore omitted from that panel.
The mechanism scores used in the main claims are reported separately as trigger F1, steering, and ablation.

\end{document}